\documentclass[journal]{IEEEtran}

\usepackage{epsfig,epstopdf}
\usepackage{cite}
\usepackage{graphicx}
\usepackage[cmex10]{amsmath}

\usepackage{array}
\usepackage{amsfonts}
\graphicspath{ {./figs/} }

\usepackage{etoolbox}
\makeatletter
\patchcmd{\@makecaption}
{\scshape}
{}
{}
{}
\makeatletter
\patchcmd{\@makecaption}
{\\}
{.\ }
{}
{}
\makeatother
\def\tablename{Table}

\usepackage{color, colortbl}
\definecolor{LightCyan}{rgb}{0.89,1,1}
\definecolor{DarkCyan}{rgb}{0.5,1,1}

\usepackage{cellspace}
\begin{document}
%
\title{Synaptic Learning with Augmented Spikes}
%
%
%

\author{
	Qiang~Yu,
	Shiming~Song, 
	Chenxiang~Ma, 
	Linqiang~Pan,\\
	Kay~Chen~Tan,~\IEEEmembership{Fellow,~IEEE}

\thanks{Q.~Yu, S.~Song and C.~Ma are with Tianjin Key Laboratory of Cognitive Computing and Application, College of Intelligence and Computing, Tianjin University, Tianjin, China.}

\thanks{L.~Pan is with Key Laboratory of Image Information Processing and Intelligent Control, Institute of Artificial Intelligence, School of Artificial Intelligence and Automation, Huazhong University of Science and Technology, Wuhan, China.}

\thanks{K.C.~Tan is with the Department of Computer Science, City University of Hong Kong, Hong Kong.}

\thanks{Corresponding author: Q.Yu (e-mail: yuqiang@tju.edu.cn).}

}

%
%

\markboth{}%
{Shell \MakeLowercase{\textit{et al.}}: Bare Demo of IEEEtran.cls for Journals}
%



\maketitle

\begin{abstract}


Traditional neuron models use analog values for information representation and computation, while all-or-nothing spikes are employed in the spiking ones. With a more brain-like processing paradigm, spiking neurons are more promising for improvements on efficiency and computational capability. They extend the computation of traditional neurons with an additional dimension of time carried by all-or-nothing spikes. Could one benefit from both the accuracy of analog values and the time-processing capability of spikes?
In this paper, we introduce a concept of \textit{augmented spikes} to carry complementary information with spike coefficients in addition to spike latencies.
New augmented spiking neuron model and synaptic learning rules are proposed to process and learn patterns of augmented spikes. We provide systematic insight into the properties and characteristics of our methods, including classification of augmented spike patterns, learning capacity, construction of causality, feature detection, robustness and applicability to practical tasks such as acoustic and visual pattern recognition.
The remarkable results highlight the effectiveness and potential merits of our methods.
Importantly, our augmented approaches are versatile and can be easily generalized to other spike-based systems, contributing to a potential development for them including neuromorphic computing.


\end{abstract}

%
\IEEEpeerreviewmaketitle

\section{Introduction}

\begin{figure}[!htb]
	\centering\includegraphics[width=0.48\textwidth]{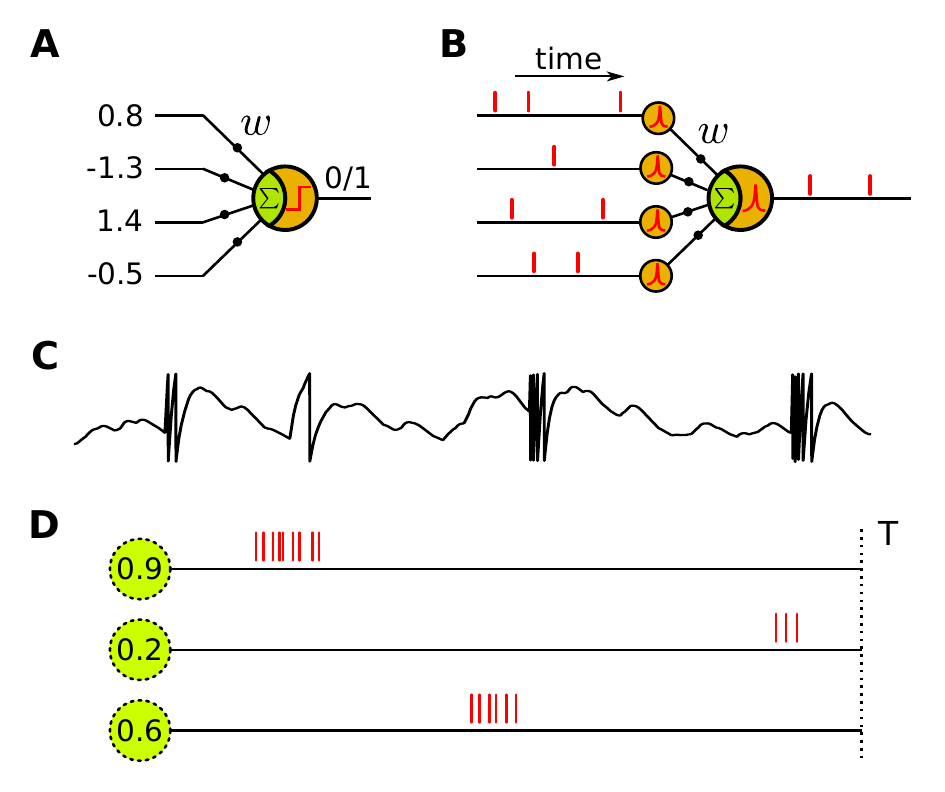}
	\caption{Schematic demonstration of neuroarchitectures and biological phenomena. \textbf{A}, a standard Perceptron neuron model. \textbf{B}, a typical spiking neuron model. \textbf{C}, a common bursting phenomenon of biological neurons in central nervous systems. \textbf{D}, a conceivable neural code utilizes both latency and spike count (inspired by \cite{gollisch2008rapid,Hopfield95}).}
	\label{Fig:motivation}
\end{figure}

\IEEEPARstart{H}{uman} brain, consisting of massive interconnected neurons, is remarkable on various cognitive capabilities such as learning, memory, decision making, etc., while retains a high power-efficiency over modern-day supercomputers \cite{kandel2000principles}.
Leveraging intuitions about operations that neurons perform in the brain, the abstract, simple and yet powerful neuron models, such as perceptron \cite{rosenblatt1958perceptron} (see Fig.~\ref{Fig:motivation}\textbf{A} for illustration), were proposed with aims to produce an approaching intelligence as compared to the brain. Incorporating more inspirations from the biological nervous systems and structures, artificial neural networks (ANNs) were proposed with simple abstractions of the biological counterparts to produce artificial intelligence to certain extent. With decades of development, ANNs, following the emergence of a family of machine learning techniques called deep learning \cite{lecun2015deep}, have achieved ubiquitous success in various fields such as face and speech recognition, which are frequently performed in our daily lives in recent years \cite{lane2017squeezing}. 
Their remarkable achievements can be at least attributed to massive data resources grabbed from the Internet and powerful computing platforms equipped with graphical processing units (GPUs).
The big data makes ANNs be generalized better, while GPUs accelerate both the training and inference of ANNs.

Despite the recent advances of ANNs, their dependency on huge amount of training data and modern high-performance computing facilities to adapt massive parameters for a given task makes them extremely expensive both in time and energy, as well as storage capacity \cite{severa2019training}. In some applications, such as on-board processing in smart platforms like mobile phones and autonomous drones, the demand for fast and efficient processing may still prohibit the running of large ANNs \cite{nawrocki2016mini}.
By contrast, the human brain performs impressive efficiency with a power budget of nearly 20 W \cite{krizhevsky2012imagenet}, let alone its powerful capabilities for cognitive computations. Therefore, neuromorphic computing efforts emerge to potentially overcome the current limitations of ANNs by carrying out brain-like processing \cite{nawrocki2016mini,feldmann2019all,roy2019towards}.

There is a fundamental difference between ANNs and nervous systems in the brain regarding to the information processing and computing paradigm. Neurons in the brain communicate and learn with discrete electrical pulses, also called spikes \cite{kandel2000principles}, while ANNs use continuous values as reflection of neurons' activations or analog of their firing rates \cite{rosenblatt1958perceptron,Hopfield95}. 
The discrete feature of spikes is believed to play an essential role in efficient and remarkable cognitive computation in the brain, and thus leads to the development of a new generation of neural networks, i.e. spiking neural networks (SNNs) \cite{maass1997networks,roy2019towards}.
Different from traditional neurons in ANNs where computations are normally degraded to spatial dimension only, spiking neurons extend their computational capability with an additional time domain (see Fig.~\ref{Fig:motivation}\textbf{B} for a demonstration). 

A typical spiking neuron receives and generates all-or-nothing (binary) spikes from upstream neurons and to downstream ones, respectively. Although it still remains unclear how information is coded exactly with spikes, there are two of the most longstanding coding schemes that are widely studied, which are the rate and temporal codes \cite{kandel2000principles,dayan2001theoretical,YuNCS,Panzeri10}.
The rate code ignores the temporal structure of the spike train, making it highly robust with respect to interspike-interval noise \cite{gerstner1997neural,london2010sensitivity}, while the temporal code has a high information-carrying capacity as a result of making full use of the temporal structure \cite{Hopfield95,Richard98,Borst99}. 
Although an increasing number of experiments have been shown in various nervous systems \cite{london2010sensitivity,kandel2000principles,gollisch2008rapid,butts2007temporal} to support different codes, it is still arguable which one dominates information coding in the brain \cite{masuda2002bridging,gutig2014spike,yu2018spike}.
Could one compound code be utilized such that it can take advantages of different codes?

It is quite often that neurons elicit spikes in a form of burst \cite{beurrier1999subthalamic,zeldenrust2018spike}: short periods of time with substantially high firing rate (see Fig.~\ref{Fig:motivation}\textbf{C} for a demonstration). This bursting phenomenon has been shown to play an important role in both reliable coding and learning \cite{naud2018sparse,simonnet2019burst,divakaruni2018long}.
Recent experimental finding in retina \cite{gollisch2008rapid} suggests that latency and spike count may serve to encode complementary stimulus features which could support a rapid and reliable analysis. 
A conceivable neural code can thus be introduced by combining the latency \cite{Hopfield95} and bursting phenomena, and we name it as latency-burst (Fig.~\ref{Fig:motivation}\textbf{D}). In this latency-burst code, the stronger the stimulus intensity, the earlier of the leading spike and the more number of spikes in the following burst. In this way, the temporal component could support rapid processing while the burst would provide complementary information for subsequent refinement.

A more radical approach can assemble spikes in the latency-burst into a single spike event, named as \textit{augmented spike} in this paper, where its appearance time carries the latency information while a multi-stage spike strength \cite{kim2018deep,chen2018fast} rather than the normal binary one can be used to reflect additional information. The concept of weighted spikes \cite{kim2018deep,chen2018fast} has shown its advantage in reduction of classification latency and the number of spikes owing to the spike strength, but the temporal benefit is rarely explored. Moreover, how neurons adapt their synaptic efficacies to extract information carried by these augmented spikes still remains very much unclear.

Learning is a vital part to almost every intelligent system. It can determine how neurons adapt their characteristics in response to external stimuli such that they can fit the environment to solve certain cognitive tasks. Spike-timing-dependent
plasticity (STDP), a timing-dependent specialization of Hebbian learning \cite{bi1998synaptic,caporale2008spike}, is one of the common characteristics in nervous systems and widely studied in computational neuroscience. A typical STDP rule is to strengthen the synaptic connection between two neurons if they have causal spike order, while to weaken it otherwise.
STDP enables neurons to process information in an unsupervised way \cite{masquelier2007unsupervised}, but with a requirement of temporal contiguity \cite{gutig06}. Many other synaptic learning rules are thus developed with different focuses \cite{gutig06,bohte02spikeprop,memmesheimer2014,wade2010swat,gutig2016}.
Learning rules like the tempotron \cite{gutig06} are developed to train neurons to have binary response of firing or not, and thus they could be used for efficient learning and classification tasks \cite{Yu2013TNN}. Other rules \cite{bohte02spikeprop,memmesheimer2014,ponulak10,YuQ2013PSD,taherkhani2018supervised} are developed to train neurons to fire at desired timings such that the temporal structure of the output could be utilized for further processing \cite{yu2016spiking,zheng2018sparse}. Other developments are proposed to train neurons to have desired number of spikes in response to given input stimuli, and such learning rules are capable of extracting embedded features and discriminating patterns encoded with different coding schemes \cite{gutig2016,yu2018spike}.
Despite various synaptic learning rules, rare one considers augmented spikes.

Although the spike-based computation and representation are promising for improvements on efficiency and computational capability \cite{roy2019towards}, 
their performance is relatively poor with respect to classification accuracy as is compared to ANNs \cite{lecun2015deep,wade2010swat,Yu2013TNN,zheng2018sparse}. Recent approaches of mapping techniques can successfully convert a trained ANN to an SNN with almost lossless accuracy \cite{cao2015spiking,diehl2015fast}, but spikes operate in a rate regime, thus leaving the efficient temporal benefit unexploited. The gap between ANNs and SNNs motivates us to combine the advantages of the fundamental elements of the both, that is the spike-based characteristic from SNNs and accurate representation in ANNs. Therefore, we heuristically introduce a new representation scheme with augmented spikes, which are also inspired and supported by biological phenomena \cite{zeldenrust2018spike,gollisch2008rapid,Hopfield95,naud2018sparse} to certain extents. However, how could neurons process and learn from these augmented spikes remains unclear. Additionally, it is also intriguing what computational benefit and application merit would one expect from learning these augmented spikes.

In this work, we propose a new spiking neuron model with the capability of processing augmented spikes. Then, several new synaptic learning rules are developed to enable neurons to learn from these augmented spikes.
The significance of our major contributions can be enumerated as follows.

\begin{itemize}
	
	\item A new spiking neuron model is proposed to process augmented spikes where additional information can be coded with spike strength in addition to latency.
	This neuron model extends the computation with an additional dimension, which could be of great significance for both processing and learning.
	
	\item We propose several synaptic learning rules that are capable of learning augmented spikes. A systematic characterization of these learning rules with respect to their computational properties is provided. These could contribute as a guideline for the development of spike-based frameworks.
		
	\item We are the first one, to the best of our knowledge, to bring the concept of augmented spike to synaptic learning. This protocol could be easily generalized to other spike-based learning and processing systems, highlighting the versatility and potential of our methods.
		
	\item We demonstrate the applicability of our methods with two practical tasks including acoustic and visual recognitions. The better performance of ours as compared to the baseline methods suggests the practical merit of our contribution.
	
\end{itemize}

The remainder of this paper is structured as follows. Section~\ref{sec:Methods} presents our proposed approaches and methods. Simulation results are presented in Section~\ref{sec:experiments}, followed by discussions in Section~\ref{sec:discuss}. Finally, a conclusion is provided in Section~\ref{sec:Conclusion}.



\section{Methods}
\label{sec:Methods}

In this section, we will firstly describe the neuron model proposed for processing augmented spikes. Then, synaptic learning rules are developed based on the neuron model. Various spike learning rules can be categorized according to neuron's output response. For example, a neuron can be trained to have binary spike response or multiple output spikes where their precise timings or total spike number is required for a desired response.
In this paper, we select tempotron \cite{gutig06}, PSD \cite{YuQ2013PSD} and TDP \cite{yu2018spike} as representatives of different types. New learning rules based on these are developed to train neurons to process augmented spikes. Other specific methods used in the simulations are described in the next section, as well as in the appendix.

\subsection{Augmented Spiking Neuron Model}

There are various popular spiking neuron models, such as Hodgkin-Huxley model \cite{hodgkin1952quantitative}, Izhikevich model \cite{izhikevich2003simple} and leaky-integrate-and-fire (LIF) model \cite{gerstner2002spiking,burkitt2006review}. They are proposed with certain levels of resembling behaviors of biological neurons. In this paper, we use the current-based leaky integrate-and-fire neuron model due to its simplicity and analytical tractability \cite{gutig2016,yu2018spike}. Notably, our scheme with augmented spikes can be easily applied to other spiking neuron models.

\begin{figure}[!htb]
	\centering\includegraphics[width=0.47\textwidth]{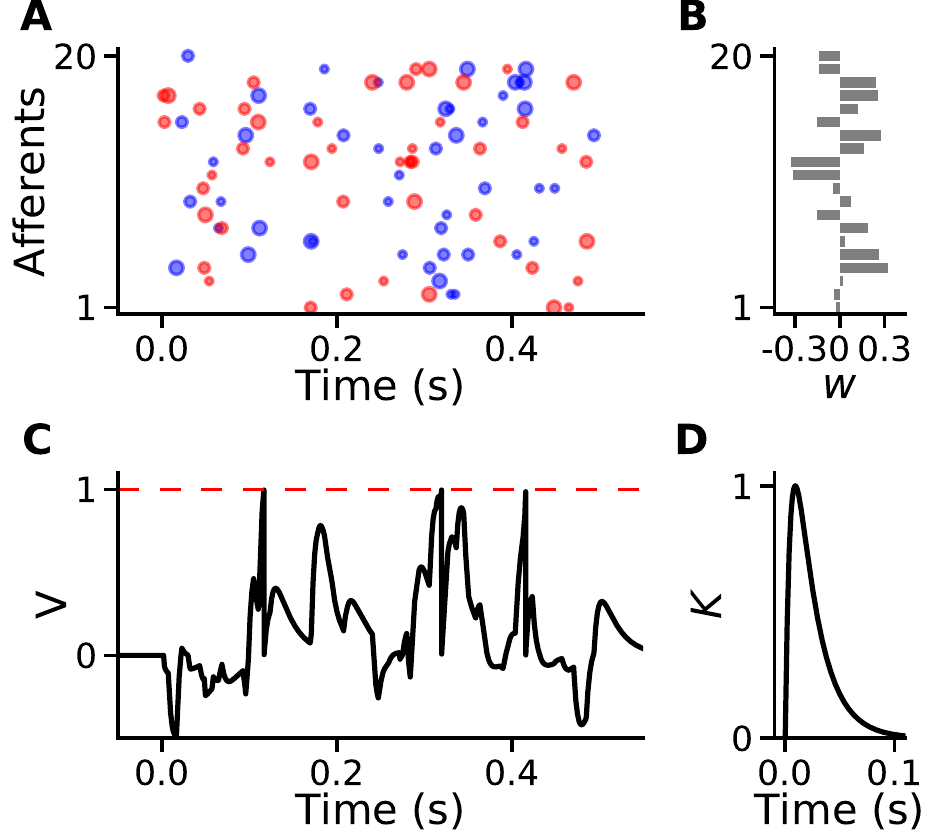}	
	\caption{Dynamics of an augmented spiking neuron model. \textbf{A}, an input pattern with augmented spikes. Each dot denotes a spike event while its color and size represent the polarity and spike strength, respectively. \textbf{B}, synaptic weights of the corresponding afferents. \textbf{C}, the resulting membrane potential of the neuron in response to the pattern in \textbf{A}. The red dashed line represents neuron's firing threshold. \textbf{D}, normalized kernel of post-synaptic potential (PSP).}
	\label{Fig:NeuronDyms}
\end{figure}

Different from a typical binary spike, the augmented one utilizes the spike strength to carry additional information, and we name this quantity as spike coefficient in the following. Spike coefficients can be any analog or discrete values carrying information of both polarity and magnitude (Fig.~\ref{Fig:NeuronDyms}\textbf{A}). In order to process augmented spikes, neurons need to incorporate the spike coefficients into their evolving dynamics.
Like a standard one, our neuron model continuously integrates afferent spikes into its membrane potential and elicit output spikes whenever a firing condition is matched. In a normal case with binary spikes, each afferent spike will result in a post-synaptic potential (PSP) whose peak is solely controlled by synaptic weight. In our augmented spiking neuron model, we additionally use spike coefficients to modulate the effect of PSPs.

The evolving dynamics of our neuron model with $N$ synaptic afferents is described as
\begin{equation}
\label{Eq:neuron}
V(t) = \sum_{i=1}^N w_i\sum_{t_i^j<t} c_i^j K(t-t_i^j) - \vartheta \sum_{t_\mathrm{s}^j<t} \exp{\left(-\frac{t-t_\mathrm{s}^j}{\tau_\mathrm{m}}\right)} 
\end{equation}
where $\vartheta$ denotes the firing threshold and $t_\mathrm{s}^j$ represents the time of the $j$-th output spike. $w$ is the synaptic weight. $t_i^j$ represents the $j$-th input spike time from the $i$-th afferent, with $c_i^j$ denoting the corresponding spike coefficient. $\tau_\mathrm{m}$ is the time constant of the neuron model. Kernel $K$ determines the shape of PSPs, and is defined as
\begin{equation}
K(t-t_i^j) = V_0\left[\exp{\left(-\frac{t-t_i^j}{\tau_\mathrm{m}}\right)}-\exp{\left(-\frac{t-t_i^j}{\tau_\mathrm{s}}\right)}\right]  
\end{equation}
where $V_0$ is a constant parameter that normalizes the peak of the kernel to unity. $\tau_\mathrm{s}$ represent the time constant of the synaptic currents.

As can be seen from Fig.~\ref{Fig:NeuronDyms}, each augmented spike will result in a PSP whose appearance is determined by the time of spike event, and its peak is controlled by both synaptic weight and spike coefficient. Whenever the neuron's membrane potential crosses its firing threshold, an output spike is elicited, followed by a reset dynamics described in Eq.~(\ref{Eq:neuron}).

\subsection{Augmented Tempotron Rule}

The tempotron (abbreviated as `Tmp' in this paper) proposed in \cite{gutig06} is an efficient rule to train neurons to make decisions by firing or not. 
Following the principles in the original tempotron rule, we propose an extension, the augmented tempotron (`AugTmp') rule, to empower neurons the ability to learn from augmented spikes.

In the AugTmp rule, the neuron can only fire a single spike due to the constraint of a shunting mechanism \cite{gutig06}, and the dynamics of which is slightly different from the one described in Eq.~(\ref{Eq:neuron}).
A neuron is required to elicit a single spike in response to a target pattern ($P^+$) and to keep silent to a null one ($P^-$). If the neuron failed to have a desired response to a given pattern, the learning will be carried out to adapt synaptic weights. A gradient descent method can be applied to minimize the cost defined by the distance between the neuron's maximal potential and its firing threshold, leading to the following AugTmp rule:
\begin{equation}
\Delta w_i=\left\{\begin{array}{rl}
\eta\sum_{t_i^j<t_{\mathrm{max}}} c_i^j K(t_{\mathrm{max}}-t_i^j),  &\mbox{ if $P^+$ error;}\\
-\eta\sum_{t_i^j<t_{\mathrm{max}}} c_i^j K(t_{\mathrm{max}}-t_i^j), &\mbox{ if $P^-$ error;}\\
0, &\mbox{ otherwise.}
\end{array}
\label{eq:tempotronRule}
\right.
\end{equation}
where $t_{\mathrm{max}}$ represents the time at the maximal potential and $\eta$ is the learning rate controlling the modification size of each step on weights.

Like the tempotron rule, decision performance of the AugTmp rule can be improved by incorporating other mechanisms such as grouping \cite{Yu2013TNN} and voting with maximal potential \cite{dennis2013temporal}. 

\subsection{Augmented PSD rule}

There is a family of learning rules \cite{bohte02spikeprop,ponulak10,yu2016spiking,florian2012chronotron} that can train the neurons not only to fire but also at exact timings as desired. Due to the efficiency and effectiveness of the PSD rule \cite{YuQ2013PSD,yu2016spiking}, we select it as a representation of this type of rules, and develop our augmented one based on it (abbreviated as `AugPSD').

The AugPSD rule is proposed to train neurons to fire at desired spike times in response to a target spike pattern. In such a way, the temporal domain of the output could be potentially utilized for information transmission as well as multi-category classification \cite{YuQ2013PSD,yu2016spiking}. 
The learning rule is implemented to minimize the difference between the desired ($t_\mathrm{d}$) and the actual ($t_\mathrm{o}$) output spike times. Following a similar derivation in \cite{YuQ2013PSD}, our AugPSD rule is thus given as:
\begin{align}
\label{Eq:AugPSD}
\Delta w_i 
&=\eta \Bigg [ \sum_g\sum_j c_i^j K(t_\mathrm{d}^g-t_i^j)H(t_\mathrm{d}^g-t_i^j) \\ \nonumber
&- \sum_h\sum_j c_i^j K(t_\mathrm{o}^h-t_i^j)H(t_\mathrm{o}^h-t_i^j) \Bigg ]
\end{align}
where $H(\cdot)$ denotes the Heaviside function.

In our AugPSD rule, a long-term depression will occur to decrease the weights when the neuron erroneously elicits an output spike, while a long-term potentiation will increase them when the neuron fails to fire at a desired time.
Different distance metrics can be developed to measure the similarity between the desired and actual output spike trains such that they could be used in both training and evaluation. For example as in \cite{YuQ2013PSD}, a distance metric can be calculated as
\begin{equation}
Dist = \frac{1}{\tau}\int_0^\infty [f(t)-g(t)]^2dt
\label{Eq:Dist}
\end{equation}
where $f(t)$ and $g(t)$ are filtered signals of the two spike trains, and $\tau$ is a time constant.

Although this metric can give a precise measurement, the choice of a critical value for termination in the training could be difficult. Additionally, the calculation of this metric would complicate the computation in a spike-based framework.

In this paper, we introduce a much simpler and efficient approach, i.e. the coincidence metric, to measure the distance. We introduce a margin parameter $\zeta$ to control the precision of the coincidence detection. The output spike time will be regarded correct if it falls into the region of $t_\mathrm{d}-\zeta \leq t_\mathrm{o} \leq t_\mathrm{d}+\zeta$. This margin parameter and coincidence detection can facilitate the learning.

\subsection{Augmented TDP rule}

Recently, a new family of learning rules are developed to train neurons to fire a certain number of spikes instead of requiring them at precise times \cite{gutig2016,yu2018spike}. 
They have shown remarkable performance as compared to others for making decision and exploring temporal features from signals due to their multi-spike characteristic. 

We adopt the TDP rule \cite{yu2018spike} in this paper, due to its efficiency and simplicity, to develop our new augmented TDP rule (abbreviated as AugTDP). 
The AugTDP rule is developed based on the spiking characteristic of multi-spike neurons, namely spike-threshold-surface (STS), which describes the relation between the neuron's output spike number and its firing threshold \cite{gutig2016}. 
A higher threshold value will normally result in a lower number of output spikes.
STS is characterized by a group of critical threshold values, $\vartheta^*_k$, that denotes the position of the neuron's firing threshold at which point its output spike number would jump around the point $k$. A neuron's actual output spike number can thus be directly reflected by STS. Therefore, modifications of the critical threshold values can result in a desired spike number. Following steps in \cite{gutig2016,yu2018spike}, the critical threshold value can be determined as
\begin{equation}
\vartheta^*=V(t^*) = U(t^*) - \vartheta^* \sum_{j=1}^m \exp{\left(-\frac{t^*-t_\mathrm{s}^j}{\tau_\mathrm{m}}\right)}
\end{equation}
Here, $U(t)=\sum_{i=1}^N w_i\sum_{t_i^j<t} c_i^j K(t-t_i^j)$, and $m$ is the total number of output spikes that occur before $t^*$ that represents the time point of $\vartheta^*$. Following a similar routine as TDP rule \cite{yu2018spike}, the derivative of $\vartheta^*$ with respect to the $i$-th synaptic efficacy can be given as
\begin{equation}
\label{Eq:derivativeTDP}
\vartheta^{*'}_i = \frac{\partial V(t^*)}{\partial w_i} 
- \sum_{j=1}^m \frac{\partial V(t^*)}{\partial t_\mathrm{s}^j} \frac{1}{\dot{V}(t_\mathrm{s}^j)}
\frac{\partial V(t_\mathrm{s}^j)}{\partial  w_i}
\end{equation}

For simplicity, we denote $t_x \in \{t_\mathrm{s}^1, t_\mathrm{s}^2,..., t_\mathrm{s}^m, t^*\}$, and thus the components in Eq.~(\ref{Eq:derivativeTDP}) can be calculated as
\begin{equation}
\frac{\partial V(t_x)}{\partial w_i} = \sum_{t_i^j<t_x}c_i^j K(t_x - t_i^j)
\end{equation}
\begin{equation}
\frac{\partial V(t^*)}{\partial t_\mathrm{s}^j} = - \frac{\vartheta^*}{\tau_\mathrm{m}}\exp(-\frac{t^*-t_\mathrm{s}^j}{\tau_\mathrm{m}})
\end{equation}
\begin{flalign}
\qquad\dot{V}(t_x)&=\frac{1}{\tau_\mathrm{s}}\sum_i w_i V_0 \sum_{t_i^j<t_x} c_i^j \exp(-\frac{t_x-t_i^j}{\tau_\mathrm{s}}) \nonumber\\ 
&- \frac{1}{\tau_\mathrm{m}}\sum_i w_i V_0\sum_{t_i^j<t_x} c_i^j \exp(-\frac{t_x-t_i^j}{\tau_\mathrm{m}}) \nonumber \\ 
&+ \frac{\vartheta^*}{\tau_\mathrm{m}} \sum_{t_\mathrm{s}^j<t_x}\exp(-\frac{t_x-t_\mathrm{s}^j}{\tau_\mathrm{m}})
\end{flalign}

Eq.~(\ref{Eq:derivativeTDP}) can be used to evaluate the derivatives $d\vartheta^*/dw$, and thus our AugTDP rule can be given as
\begin{equation}
\label{Eq:AugTDP}
\Delta w =
\begin{cases}
\eta \frac{d\vartheta^*_{n_o+1}}{d w},  & \quad \text{if } n_\mathrm{o}<n_\mathrm{d};\\[4pt]
-\eta \frac{d\vartheta^*_{n_o}}{d w}  ,     & \quad \text{if } n_\mathrm{o}>n_\mathrm{d}; \\[4pt]
0  ,     & \quad \text{otherwise.}  \\
\end{cases}
\end{equation}
where $n_\mathrm{d}$ and $n_\mathrm{o}$ represent the desired and actual output spike numbers, respectively. 
The principle of this rule is to decrease (increase) the corresponding $\vartheta^*$ that is greater (smaller) than $\vartheta$ with an LTD (LTP) process if the neuron fails to have a desired response.

\section{Simulation Results}
\label{sec:experiments}

In this section, we provide systematic evaluations on our proposed learning rules, including classification of spatiotemporal spike patterns, learning capacity, construction of causality, feature detection and discrimination, robustness and applicability to practical tasks. Some experimental setups are described in each task while others are given in the Appendix section. In all experiments, the default number of afferents is set as $N=500$ except for the acoustic and visual recognition tasks where it is determined by the encoding layer.

\subsection{Augmented Tempotron}

In this experiment, we study the capability of our AugTmp rule on classifying the spatiotemporal spike patterns where augmented spikes are used for information representation. We construct random patterns with each afferent firing at a Poisson rate of 2 Hz over a time window of $T=0.5$ s. In addition to the randomness on the timing of each spike, the spike coefficient of each spike event is also randomly selected, e.g. from the set of \{0.5, 1, 1.5\} with equal possibility in this experiment.

\begin{figure}[!htb]
	\centering\includegraphics[width=0.45\textwidth]{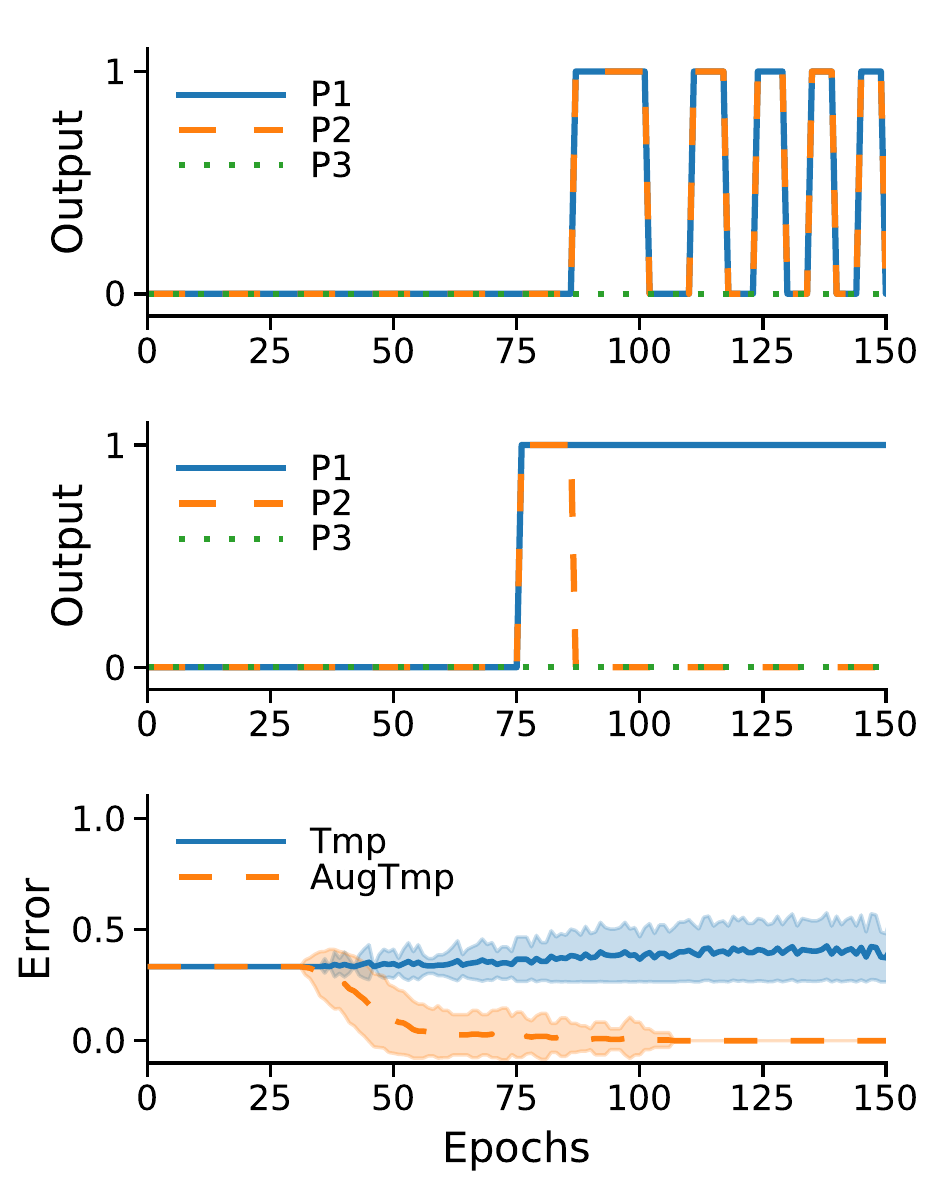}
	\caption{Augmented spike pattern classification (P1-3) with the Tmp and AugTmp rules.  The top and middle panels show a single learning run of the Tmp and AugTmp rules, respectively. The bottom panel is the learning performance averaged over 100 runs.}
	\label{Fig:augtmp}
\end{figure}

We generate three patterns, i.e. P1-3, in this task. Both P1 and P3 are independently constructed with both their spike timings and coefficients being randomly generated. P2 shares the same spike timings with P1 but with different random spike coefficients. We use both Tmp and AugTmp rules to train the neuron to elicit an output spike in response to P1 while remain silent to the other two.

As can be seen from Fig.~\ref{Fig:augtmp}, the Tmp rule cannot separate all the patterns while our AugTmp rule can successfully classify all of them as expected. This is because that there is no difference between P1 and P2 to the Tmp rule as it cannot process the augmented part, i.e. spike coefficients, thus resulting in an error rate around one third. Differently, our AugTmp rule can utilize both the spike timings and coefficients to recognize patterns, suggesting the effectiveness of our augmented scheme.

\subsection{Learning Capacity}

How many patterns could a neuron discriminate? A useful measurement of the load can be calculated as the number of patterns $p$ relative to the number of weights, denoted by $\alpha=p/N$. An important characteristic of the neuron's capacity is the maximal load it can learn \cite{gutig06}, and we denote it as the critical capacity $\alpha_c$. In this experiment, we study how many patterns a neuron can learn from augmented spikes with our learning rule.

\begin{figure}[!htb]
	\centering\includegraphics[width=0.45\textwidth]{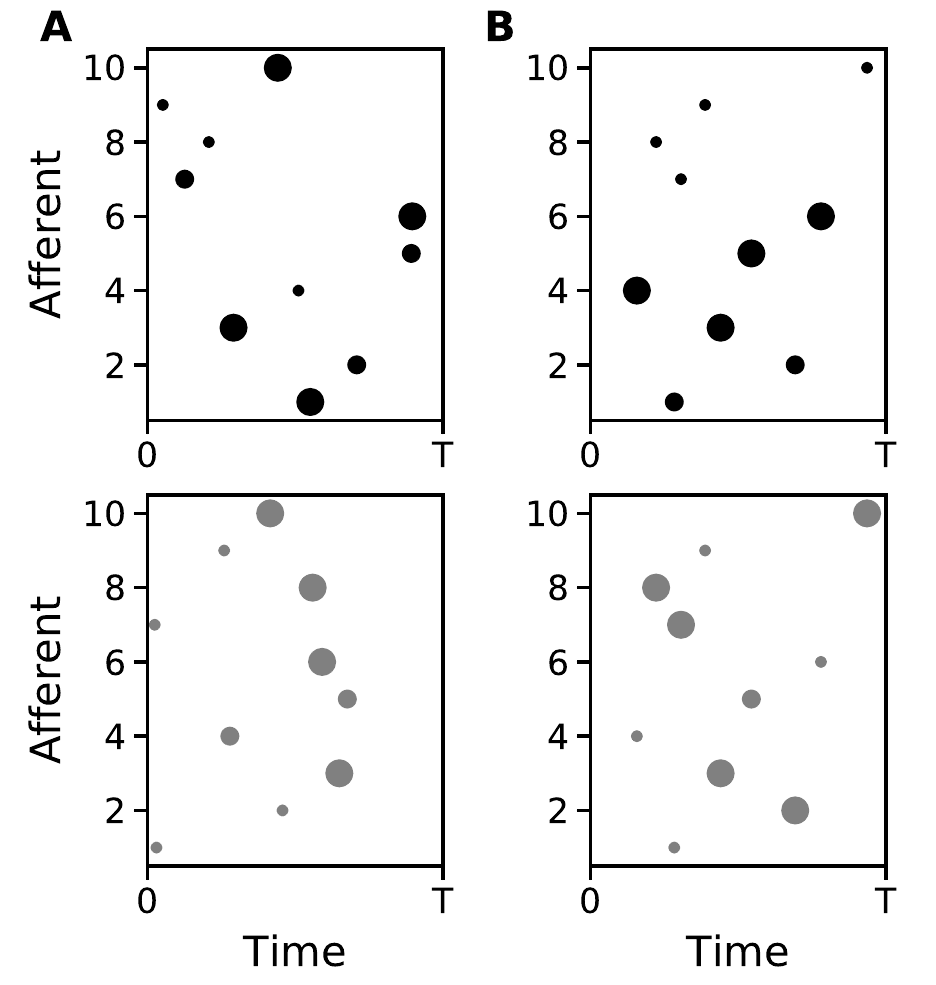}
	\caption{Illustration of patterns with augmented spikes used in the capacity task. Each dot denotes an augmented spike with its size reflecting the spike coefficient. \textbf{A}, both the spike timing and coefficient are randomly and independently generated in each pattern. \textbf{B}, the target (black) and null (gray) categories share the same spike timings while spike coefficients are randomly generated for each pattern.}
	\label{Fig:cpt_ptn}
\end{figure}

\begin{figure}[!htb]
	\centering\includegraphics[width=0.45\textwidth]{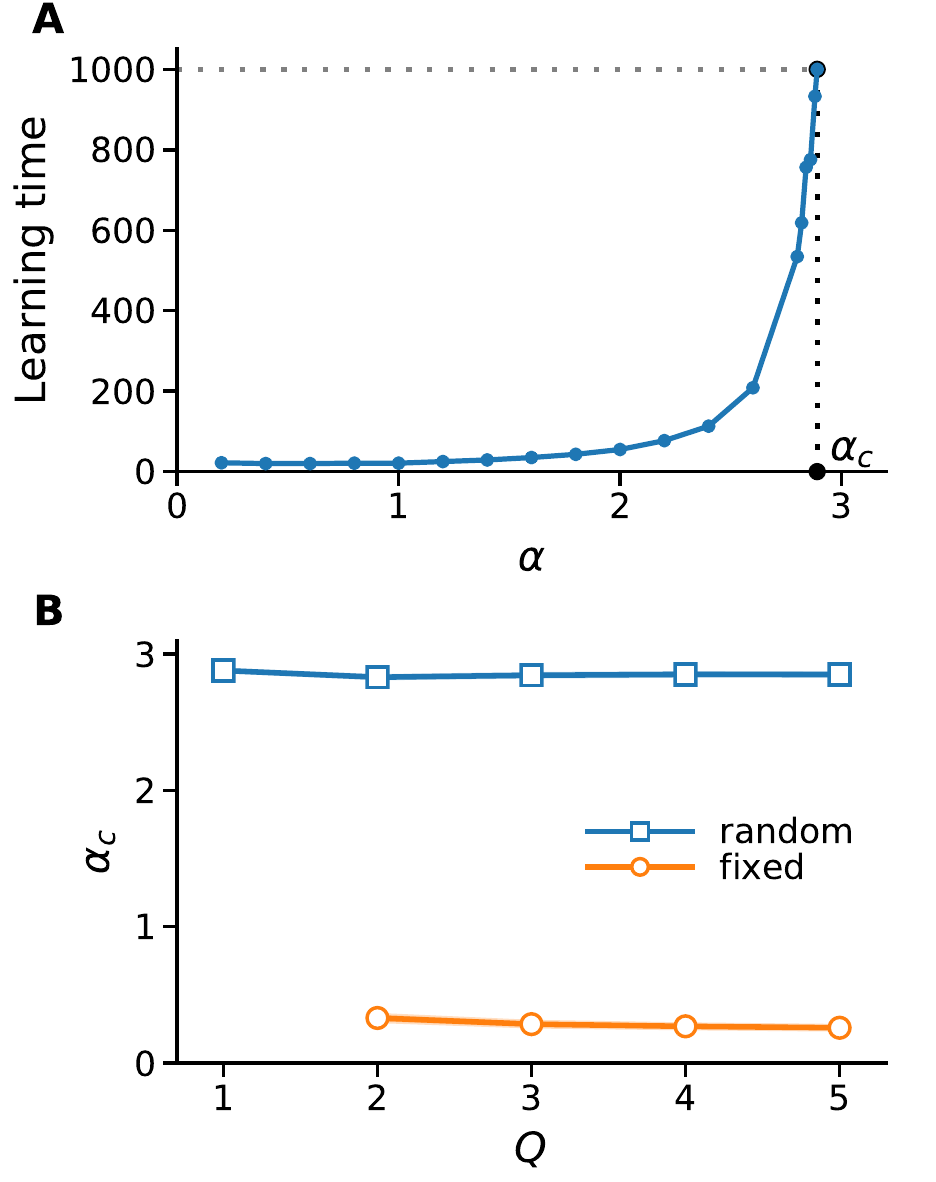}
	\caption{Capacity with augmented spikes. \textbf{A}, demonstration of a capacity curve ($Q=3$), i.e. learning time versus $\alpha$. Highlighted $\alpha_c$ denotes our approach to quantify the critical capacity. \textbf{B}, capacity performance for both random and fixed spike latencies. Data were averaged over 20 runs.}
	\label{Fig:cpt_ac}
\end{figure}

Similar to setups in \cite{gutig06}, we assess the performance of our AugTmp rule in classifying populations of single spike latencies. In each pattern, each afferent contains a single spike that occurs at a time randomly chosen from a uniform distribution over a time window $T=0.5$ s. In addition to random latencies, we assign random discrete values to spike coefficients. 
For simplicity, we generate $Q$ possible numbers that are evenly spaced over the range between 0.5 and 1.5, and then each spike coefficient is randomly selected from these $Q$ values with equal possibility. Notably, $Q=1$ will result in a single coefficient of 1, which returns to the normal binary spike patterns.
The task consists of $p$ patterns, each of which is randomly assigned to the target or null classes with equal possibility. The neuron is required to fire in response to the target class while remain silent otherwise.

We study two scenarios for the capacity task: random and fixed latencies (see Fig.~\ref{Fig:cpt_ptn}). In the random-latency case, both spike timing and coefficient of each spike in each pattern are randomly generated, while patterns share the same latencies but with different random spike coefficients in the fixed-latency case. The neuron is trained to successfully classify all patterns, and the corresponding cycles at convergence is recorded as the learning time. The appearance of all $p$ patterns during training is denoted as one cycle. 

In order to have a reliable measurement of the learning time, we use the median, rather than the mean, of 100 single evaluations due to the heavy-tail effect which could skew it by a small portion of extremely large values (data were not shown).
We obtain $\alpha_c$ from the capacity curve (Fig.~\ref{Fig:cpt_ac}\textbf{A}) at the point it crosses a learning time of 1000. Normally, the more patterns to learn, the longer of the convergent time. We choose our selection with a consideration of computational cost.

Fig.~\ref{Fig:cpt_ac}\textbf{B} shows our capacity performance. For both random and fixed latencies, the capacity is almost invariant to the level of discretization, $Q$, on the spike coefficient. This is because the spiking neuron could be approximated by a two-layer perceptron whose capacity is invariant to the input statistics \cite{gutig06}. In the random-latency case, the capacity is around 2.9 which is consistent to that in the tempotron \cite{gutig06}. Our result indicates neurons cannot store more patterns with augmented spikes. If we remove the randomness in the latency, the capacity drops to a value around 0.3, indicating the importance of the temporal domain for information processing in SNNs. Notably, the capacity for normal binary spikes ($Q=1$) in the fixed-latency case is 0 (data not shown), indicating the ability of our learning to extract information from spike coefficients.

\subsection{Augmented PSD}

In this experiment, we examine the ability of our AugPSD rule to train neurons to fire at desired times in response to a given single-spike latency pattern. In the spike pattern, each afferent fires a single spike at a time point that is randomly chosen within the time window $T=0.3$ s. We impose spike coefficients of 2.0, 1.0 and 0.5 to spikes falling in the region of [0, 0.1), [0.1, 0.2) and [0.2, 0.3], respectively. The neuron is trained to fire at 0.1 and 0.2 s with a margin precision of $\zeta=1$ ms in response to the pattern.

\begin{figure}[!htb]
	\centering\includegraphics[width=0.45\textwidth]{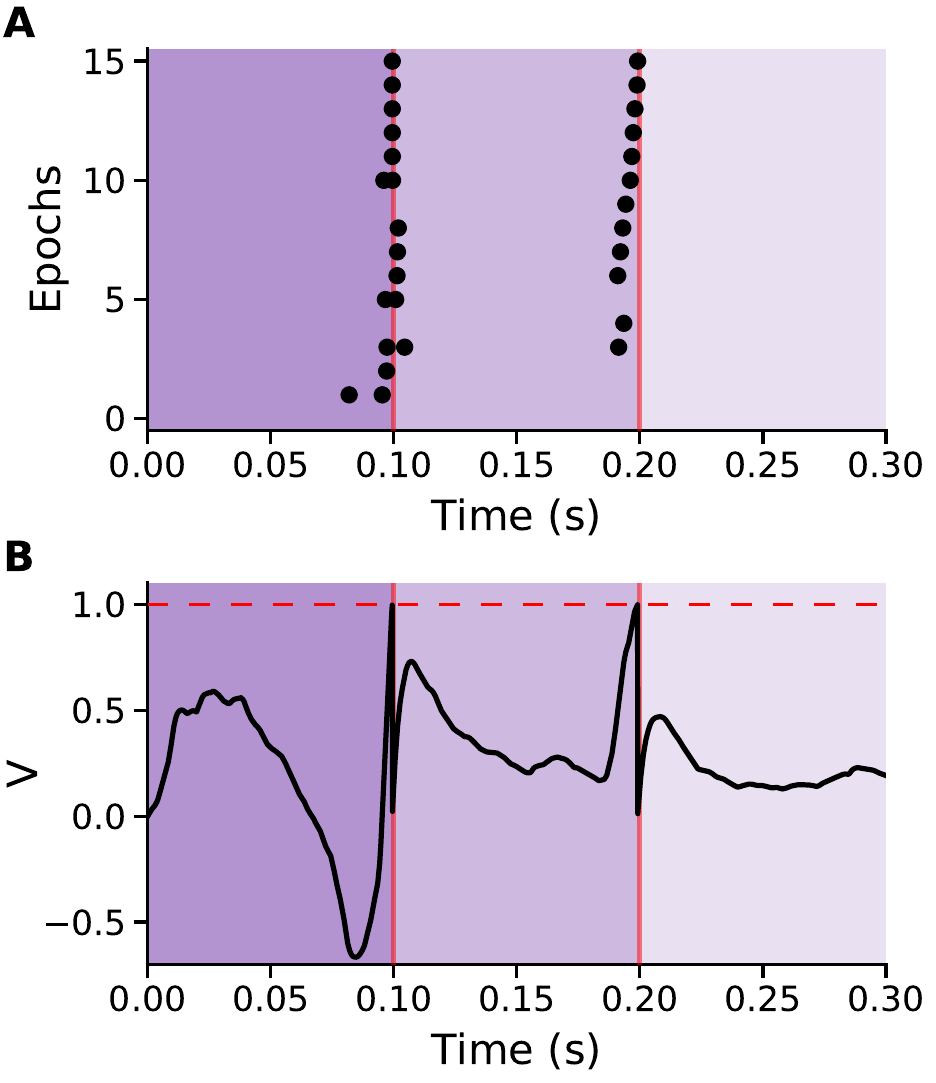}
	\caption{Learn to fire at desired times with the AugPSD rule. Vertical red bars denote the region for desired times, and shaded purple areas represent regions with spike coefficients of 2.0, 1.0 and 0.5 (from dark to light). \textbf{A}, a typical learning run with each dot representing an output spike. \textbf{B}, membrane dynamics of the neuron after learning. Dashed red line represents the firing threshold.}
	\label{Fig:psd_demo}
\end{figure}

As can be seen in Fig.~\ref{Fig:psd_demo}, our AugPSD rule can successfully train the neuron to fire at desired positions within around 10 training epochs, suggesting the effectiveness and efficiency of our rule.

\begin{figure}[!htb]
	\centering\includegraphics[width=0.45\textwidth]{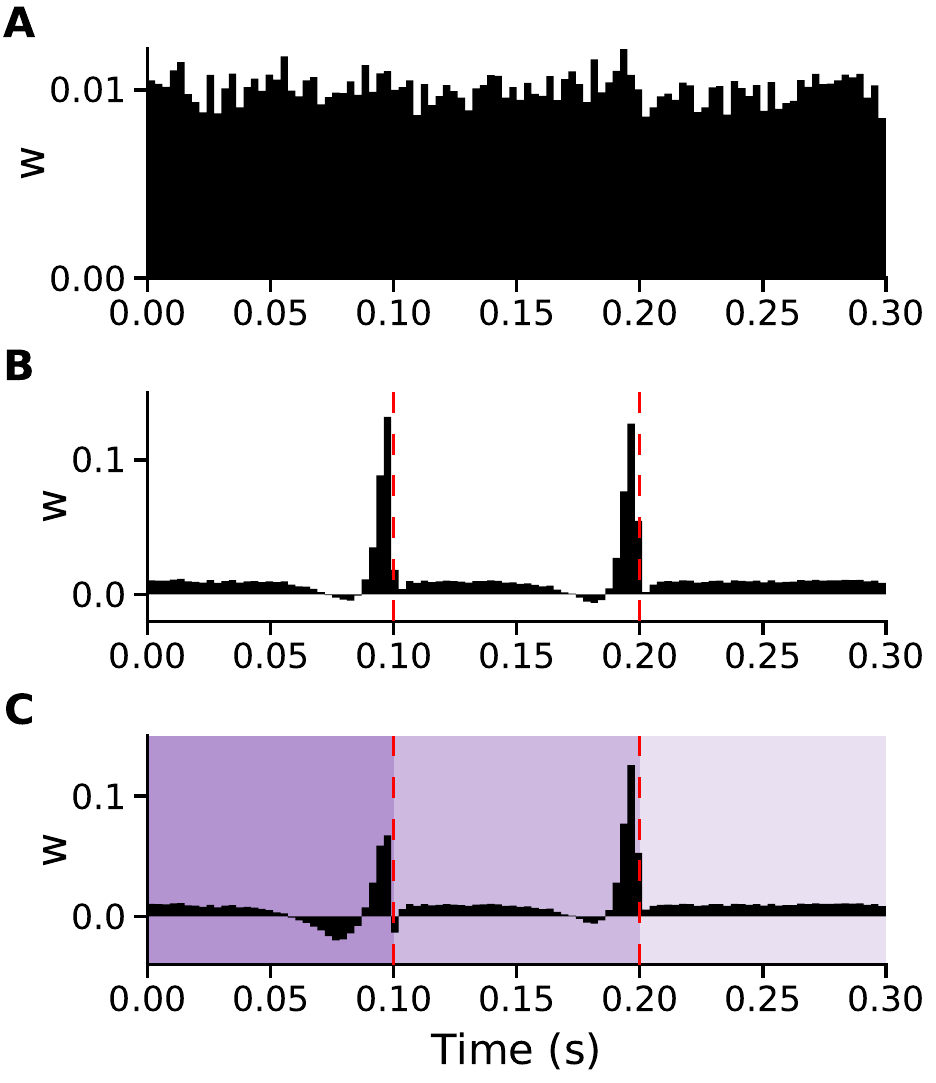}
	\caption{Causality. \textbf{A}, initial distribution of synaptic weights. \textbf{B} and \textbf{C} are weight distributions after learning with the PSD and AugPSD rules, respectively. Red dashed lines denote the position of desired times. Synaptic weights are aligned according to the chronological order of their spike times. Data were collected over 100 runs.}
	\label{Fig:psd_wdist}
\end{figure}

Moreover, our AugPSD rule is capable of constructing causal connections based on the augmented spikes. Fig.~\ref{Fig:psd_wdist} shows that the PSD rule is incapable of processing spike coefficients, and thus result in a similar distribution around the two desired times (Fig.~\ref{Fig:psd_wdist}\textbf{\textbf{B}}). This is because each spike is treated as the same in the PSD rule. Differently, our AugPSD rule is sensitive to spike coefficients and thus build causal connections based on it. 
The resulted peaks of causal connections before desired times are reversely proportional to the spike coefficients of their causal inputs (Fig.~\ref{Fig:psd_wdist}\textbf{C}). This reflects the capability of our AugPSD rule to extract information embedded in spike coefficients.

\subsection{Augmented TDP}

In this experiment, we evaluate the ability of our AugTDP in detecting and discriminating features embedded in a background. 

\begin{figure}[!htb]
	\centering\includegraphics[width=0.45\textwidth]{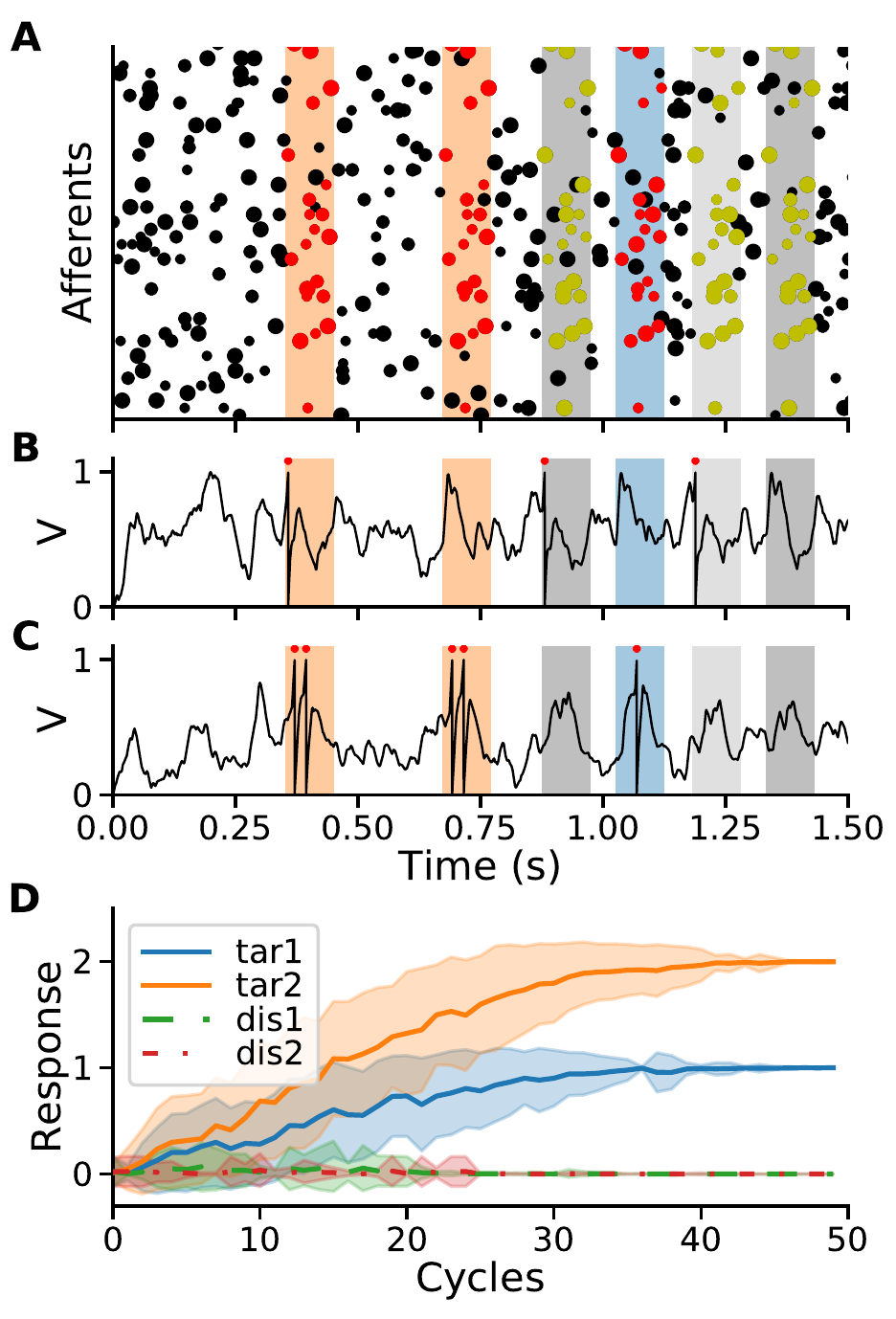}
	\caption{Feature detection and discrimination with AugTDP. \textbf{A}, a demonstration of a trial spike pattern with only 10\% afferents being plotted for a better visualization. Each dot denotes an augmented spike from background (black) or feature patterns (color). Color and gray shaded areas represent positions of the target and distractor features, respectively. \textbf{B} and \textbf{C} show the neuron's dynamics after learning with TDP and AugTDP, respectively. Each red dot denotes an output spike. \textbf{D}, the evolving trace of neurons with AugTDP in response to the target (`tar1,2') and distractor (`dis1,2') features. Data were averaged over 50 runs.}
	\label{Fig:lfeature}
\end{figure}

Similar to experimental setups in \cite{gutig2016,yu2018spike}, we construct four feature patterns with a Poisson firing rate of 4 Hz over a time window of 0.1 s. These features are randomly embedded with a mean Poisson appearance time of 3 in a background of $T=2$ s which has a same firing statistics as the features. Then a global noise with a Poisson rate of 1 Hz is imposed on the pattern. Differently, each spike contains a coefficient value that is randomly selected from the set of \{0.5, 1.0, 1.5\} with equal possibility. To make the task more challenging, we force all feature patterns share the same spike timings. This means their difference only exists in the spike coefficients. The neuron is required to fire 2 and 1 spikes in response to two randomly selected features (target), while to remain silent to others (distractor and background).

Fig.~\ref{Fig:lfeature} shows the learning performance for this task. The TDP rule fails to detect and discriminate the target features from the others (Fig.~\ref{Fig:lfeature}\textbf{B}). This is because TDP cannot process spike coefficients and thus it is incapable to discriminate feature patterns as their difference only exists in the coefficients. Thanks to the augmented capability for processing augmented spikes, our AugTDP can successfully learn to have desired behavior in response to the pattern within tens of training cycles (Fig.~\ref{Fig:lfeature}\textbf{C} and \textbf{D}).

\subsection{Robustness}

In this task, we study the robustness performance of our augmented learning rules in multi-category classification tasks. We randomly generate three spike patterns with their spike timings following a Poisson rate of 2 Hz over $T=0.5$ s and their spike coefficients being randomly selected from the set of \{0.5, 1.0, 1.5\}. These three patterns are then fixed as the template of each category. We train three neurons to learn these categories with each one corresponding to one category as its target. For AugTmp, the neurons are required to fire at least one spike in response to their target categories while remain silent otherwise. For both AugPSD and AugTDP, neurons are trained to fire 10 spikes in response to their targets otherwise to remain silent. The desired timings for AugPSD are evenly spaced over $T$ with a precision of $\zeta=10$ ms.

\begin{figure}[!htb]
	\centering\includegraphics[width=0.47\textwidth]{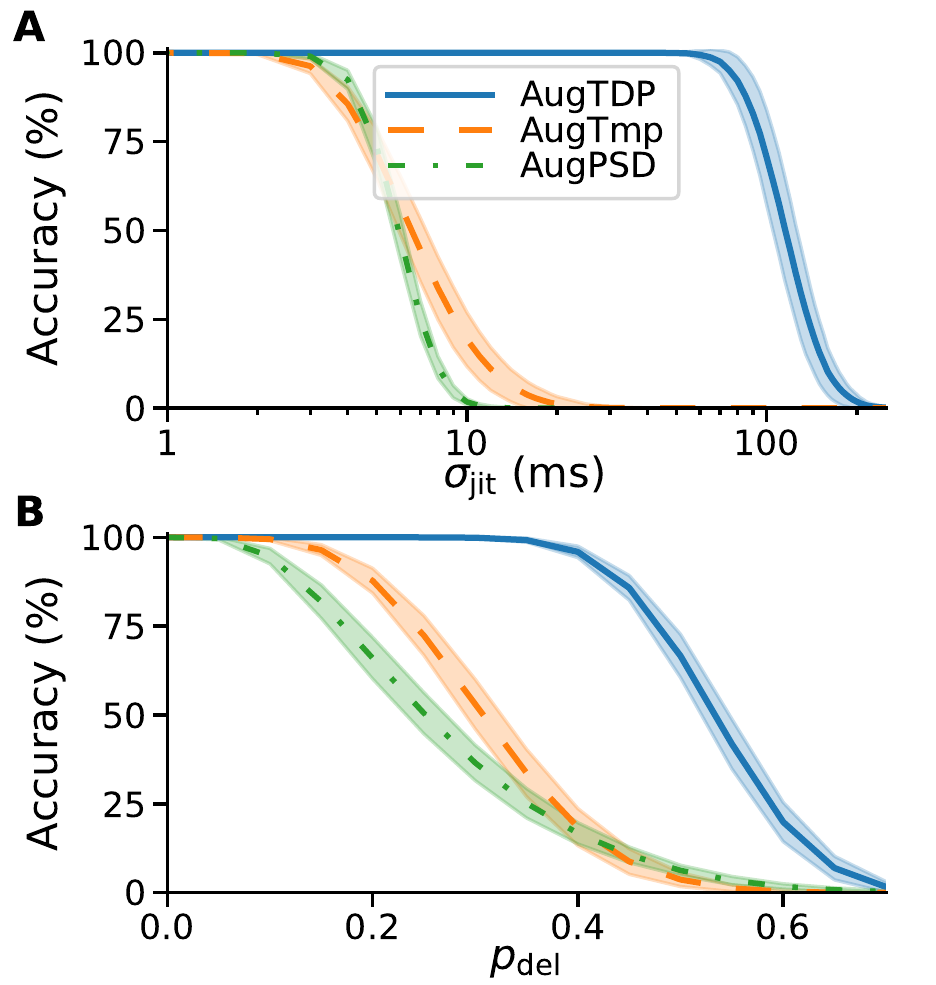}
	\caption{Robustness of different augmented learning rules under scenarios of spike jitter noise $\sigma_\mathrm{jit}$ (\textbf{A}) and random spike deletion $p_\mathrm{del}$ (\textbf{B}). Data were averaged over 100 runs.}
	\label{Fig:robustness}
\end{figure}

Neurons are trained and evaluated under two noise cases: spike jitter noise $\sigma_\mathrm{jit}$ and spike deletion noise $p_\mathrm{del}$. In the first case, the timing of each spike is jittered with a Gaussian distribution with mean of 0 and standard deviation of $\sigma_\mathrm{jit}$. In the second case, each spike is randomly deleted with a probability of $p_\mathrm{del}$. During training, we use $\sigma_\mathrm{jit}=2$ ms and $p_\mathrm{del}=0.1$ for each case.
In the evaluation phase, we use a strict readout scheme for different rules to provide clear insight into their robustness. For AugTmp, a pattern is regarded as correctly classified only if its corresponding neuron fires. For multi-spike rules of AugPSD and AugTDP, we use half of the desired spike number used in the training as the readout indicator in the evaluation. That is, only if the corresponding neuron elicits more than 5 spikes, the pattern is treated as correctly recognized.

Fig.~\ref{Fig:robustness} shows the performance of our learning rules. AugTDP is the most robust one for both noise cases as compared to the others. It can tolerate up to 80 ms jitter noise and 40\% deletion while still remains at a high accuracy over 80\%. This suggests the strength of multi-spike learning and readout. For AugPSD, neurons are constrained to fire at desired times thus limiting their ability to fully explore features over the whole time window. The bottleneck for AugTmp is its binary response of firing or not, which restricts neurons to make decisions based on only one temporal feature. Notably, advanced techniques \cite{yu2016spiking,dennis2013temporal,yu2019robust} can be applied to improve the performance of AugTmp and AugPSD, but they would complicate the decision procedure in a spike-based framework.

\subsection{Acoustic Pattern Recognition}

\begin{figure}[!htb]
	\centering\includegraphics[width=0.47\textwidth]{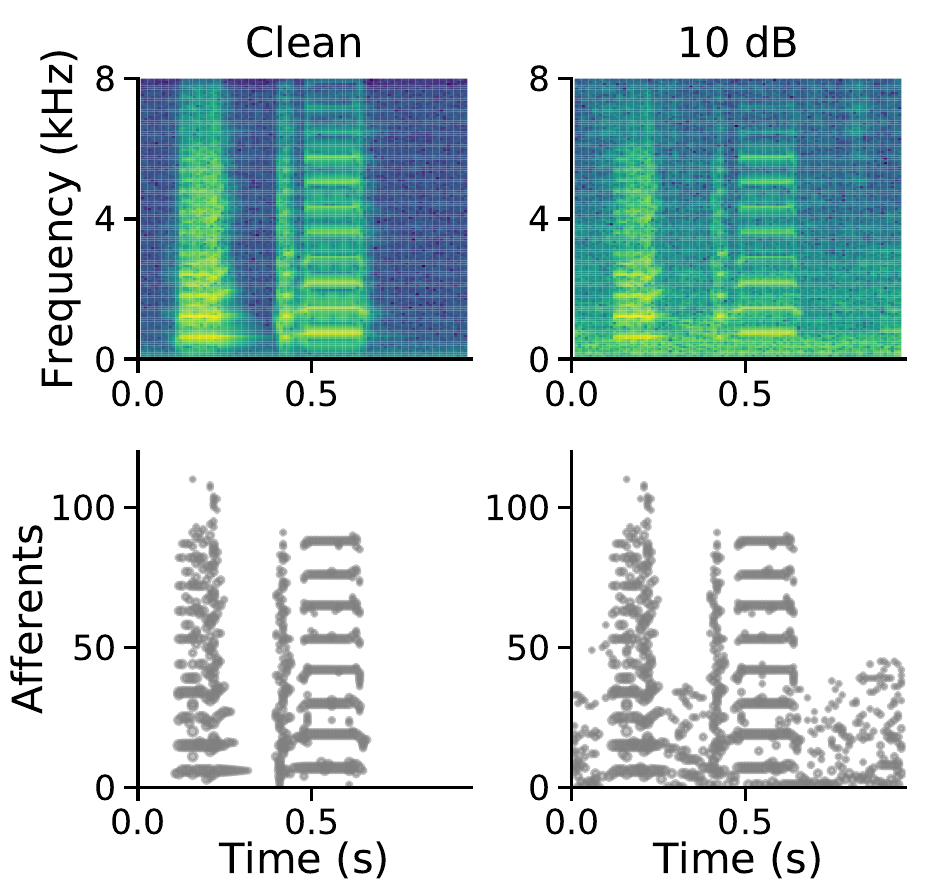}
	\caption{Demonstration of the sound encoding with augmented spikes. Top row panels are sound samples under a clean and noise of 10 dB, respectively. Bottom panels are the encoded spike patterns with the scheme in \cite{yu2019robust} but being extended to carry intensity in spikes to compose augmented ones. Each dot represents a spike with its size denoting the intensity.}
	\label{Fig:kp_encoding}
\end{figure}

Recent advances of spike-based frameworks for acoustic processing emerge with remarkable performances \cite{dennis2013temporal,yu2019robust,wu2018spiking,xiao2018spike}. In this experiment, we study the performance of our augmented rules with a more practical task, a sound recognition task used in \cite{yu2019robust}. We use the same experimental setups (details can be found in \cite{yu2019robust}) to have a better benchmark on the benefits brought by our augmented schemes.

Differently, we extend the encoding scheme of \cite{yu2019robust} with the ability to carry intensity with our augmented spikes (Fig.~\ref{Fig:kp_encoding}). In addition to the timing of each spike, its coefficient is used to carry the intensity information which is expected to facilitate the learning.

\begin{table}[!tbh]
	\centering
	\caption{Classification accuracy (in percentage \%) under mismatched condition. Shaded areas represent our proposed methods, and the bold digits denote the best across each column. `Avg$.$' indicates the average value of different conditions. Data were collected over 10 runs.}	
	\begin{tabular}{Sc||Sc Sc Sc Sc| Sc}
		\hline
		\textit{Methods}  & Clean & 20dB & 10dB & 0dB &   Avg$.$ \\ \hline 
		
		MFCC-HMM \cite{dennis2013temporal}& 99.0  & 62.1  & 34.4 & 21.8  &  54.33 \\

		SPEC-MLP \cite{yu2019robust}    & \textbf{100}  & 94.38  & 71.8 & 42.68  &   77.22 \\
		
		SPEC-CNN \cite{yu2019robust}   & 99.83  & 99.88  & 98.93 & 83.65  &  95.57 \\		
		
		SOM-SNN\cite{wu2018spiking}& 99.6  & 79.15  & 36.25 & 26.5  &  60.38 \\		
		
		LSF-SNN\cite{dennis2013temporal} & 98.5  & 98.0  & 95.3 & 90.2  &  95.5 \\
		
		LTF-SNN\cite{xiao2018spike}& \textbf{100}  & 99.6  & 99.3 & 96.2  &  98.77 \\
		
		KP-Tmp \cite{yu2019robust}  & 99.35  & 96.58  & 94.0 & 90.35  &  95.07 \\ 

		\rowcolor{DarkCyan}
		KP-AugTmp   & 99.10  & 99.20  & 98.43 & 94.63  &  97.84 \\ 
		KP-TDP \cite{yu2019robust}  & \textbf{100}  & 99.5  & 98.68 & \textbf{98.1}  &  99.07  \\	
		
		\rowcolor{DarkCyan}
		KP-AugTDP   & \textbf{100}  & \textbf{99.95}  & \textbf{99.55} & 98.03  &  \textbf{99.38}  \\
		\hline
	\end{tabular}
	\label{tab:mismatch}
\end{table}

The performance is evaluated under a mismatched condition, where training is performed with only clean samples while testing is evaluated with different levels of noises.
Table~\ref{tab:mismatch} shows the performance on the sound recognition task.
Both spike-based and conventional approaches are considered. Traditional approaches include MFCC-HMM, SPEC-MLP and SPEC-CNN. 
As a typical framework widely used in acoustic processing, MFCC-HMM performs well in clean condition, but drops rapidly with increasing noise. Deep learning techniques, SPEC-MLP and SPEC-CNN, demonstrate improvements as is compared to MFCC-HMM, but their dependency on GPUs would make them resource consuming and computationally inefficient. Thus, the spike-based approach provides an alternative possibility. Our framework with the AugTDP achieves the best performance as is compared to the other baselines. Our comparisons between close approaches with or without augmented scheme clearly highlight the corresponding improvements of our methods over the baseline ones. This indicates the additional information carried by spike coefficients, together with our augmented learning, can effectively improve the performance.

\begin{figure}[!htb]
	\centering\includegraphics[width=0.48\textwidth]{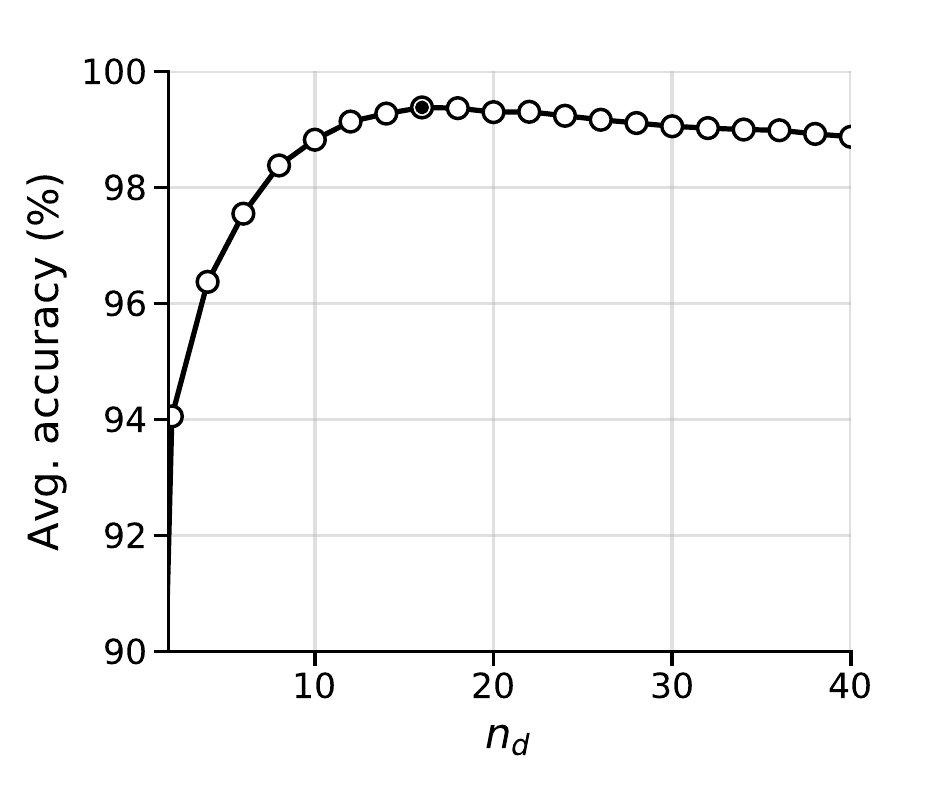}
	\caption{Sound recognition accuracy versus $n_d$. Filled circle denotes the result used in Table~\ref{tab:mismatch}.}
	\label{Fig:sound_nd}
\end{figure}

In addition, we evaluate the performance of our AugTDP with different desired spike numbers, $n_d$, used for training. Fig.~\ref{Fig:sound_nd} shows the effects of $n_d$ on the performance. For small values of $n_d$, the bigger the desired spike number, the better the recognition accuracy. This is because more spikes can be useful to extract important temporal features for a decision. However, further increase of $n_d$ will degrade the performance since the temporal interference \cite{yu2016spiking} is getting severe and thus affects the performance.

\begin{figure}[!htb]
	\centering\includegraphics[width=0.47\textwidth]{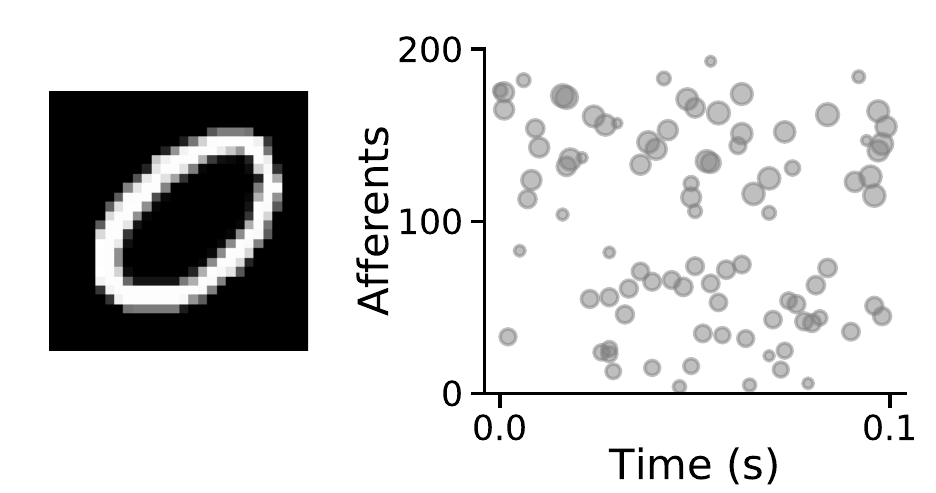}
	\caption{Demonstration of the image encoding with augmented spikes. Left, an image sample of digit 0. Right, the encoded pattern.}
	\label{Fig:img_encoding}
\end{figure}

\begin{table*}
	\caption{Test accuracies (\%) of SNNs on MNIST. Shaded areas denote our methods.}
	\begin{center}
		\begin{tabular}{Sc||Sc Sc Sc| Sc}
			\hline
			Methods  & Coding Scheme & Neurons(Structure) & \#Spikes &  Accuracy \\
			\hline
			S1C1-SNN \cite{Yu2013TNN} & Temporal &  200+10  & 200 & 78.00 \\
			
			\rowcolor{LightCyan}
			\textbf{S1C1+AugTmp} & \textbf{Temporal} & \textbf{200+10}  & \textbf{83} &  \textbf{86.10} \\
			
			\rowcolor{LightCyan}
			\textbf{S1C1+AugTDP}  & \textbf{Temporal} & \textbf{200+10} & \textbf{83} &  \textbf{86.60} \\
			
			CSNN \cite{xu2018csnn}  & Temporal &  800+400& 800  & 88.00 \\
			
			\rowcolor{LightCyan}
			\textbf{CNN+AugTmp} & \textbf{Temporal} & \textbf{800+10}  & \textbf{263} & \textbf{97.00} \\
			
			\rowcolor{LightCyan}
			\textbf{CNN+AugTDP} & \textbf{Temporal}  & \textbf{800+10}  & \textbf{263} & \textbf{97.80} \\
			
			\rowcolor{LightCyan}
			\textbf{HMAX+AugTmp}   & \textbf{Temporal} & \textbf{800+10} & \textbf{399}  & \textbf{96.90} \\
			
			\rowcolor{LightCyan}
			\textbf{HMAX+AugTDP}  & \textbf{Temporal}  & \textbf{800+10} & \textbf{399}  & \textbf{97.90} \\
			
			Dendritic Neurons \cite{hussain2014improved}  & Rate &  5000+10 & - & 90.26 \\
			
			Spike-DBN \cite{o2013real}  & Rate & 784+500+500+10 & - & 94.10 \\
			
			Spiking RBM \cite{merolla2011digital} & Rate  & 6470+1010  & - & 94.09 \\
			
			Unsupervised STDP \cite{diehl2015unsupervised} & Rate  & 784+6400 & - & 95.00 \\
			
			SNN-WS \cite{kim2018deep} & Rate(Weight) & 784+1200+1200+10 & $ 8\times10^{6} $ & 98.60 \\
			
			Weight norm \cite{diehl2015fast}  & Rate & 784+1200+1200+10  & $ 1\times10^{6} $ & 98.60 \\
			
			
			\hline
		\end{tabular}
		\label{tab:mnist}
	\end{center}
\end{table*}

\subsection{Visual Pattern Recognition}

In this section, we explore the ability of our augmented rules with a visual recognition task on MNIST dataset. Three encoding methods, namely S1C1 \cite{yu2013rapid}, HMAX \cite{riesenhuber1999hierarchical} and CNN \cite{xu2018csnn}, that convert images into spike patterns are adopted to test the performance of our augmented schemes.  The experimental setups of S1C1 and CNN are the same as \cite{yu2013rapid} and \cite{xu2018csnn}, respectively. For HMAX, we adopt the first two layers in \cite{masquelier2007unsupervised}.

In the above encoding methods, neurons will integrate information from their receptive fields and elicit spikes whose timings linearly depend on their activation values with stronger ones resulting in earlier spikes.
Differently, in this paper, we introduce a more simple and effective approach to generate afferent latencies. If the activation value of an encoding neuron crosses its firing threshold, it will fire an augmented spike at a fixed time randomly chosen from a uniform distribution between 0 and the time window $T=100$ ms, and with its activation value being assigned to the spike coefficient. 
Fig.~\ref{Fig:img_encoding} demonstrates an encoding example of our method. Each dot denotes an augmented spike and its size represents the activation value of the encoding neuron. 

Table \ref{tab:mnist} shows the test accuracies of our augmented approaches, as well as other SNN models, on the MNIST task. Firstly, we conduct a close comparison with S1C1-SNN \cite{yu2013rapid} and CSNN \cite{xu2018csnn} which use the same encoding scheme as ours but differently with normal binary spikes.
As can be seen in Table \ref{tab:mnist}, our augmented frameworks perform better than those baselines with higher accuracies, lighter network structures and fewer number of spikes. This highlights the advantages of our augmented rules for improving the performance of current spike-based frameworks. 
Besides, we also compare ours with some rate-based ones that adopt firing rates to represent information. 
The accuracy of ours still outperforms most of them, while is slightly below the SNN-WS \cite{kim2018deep} and WeightNorm \cite{diehl2015fast} which are representative approaches of mapping the weights of a trained ANN to an SNN. 
However, the high accuracy of these mapping approaches \cite{kim2018deep,diehl2015fast} comes with costs of a massive number of spikes and a complex network structure.
In contrast, our temporal-based frameworks provide efficient and effective alternatives with fewer spikes and smaller number of neurons.

\section{Discussions}
\label{sec:discuss}

Traditional ANNs distribute computations over a spatial dimension, while SNNs extend them with an additional time domain. This spatiotemporal processing capability makes spiking neurons as a new generation of neuronal models that are computationally more powerful \cite{maass1997networks,Hopfield95,Richard98,gutig06}.
One of the main differences between the traditional and spiking models is the way how information is represented: traditional ones use analog values while all-or-nothing pulses are used in spiking ones.
The spike-based computation and representation are promising for efficiency and computational capability \cite{roy2019towards}, while the non-spike based ones are widely applied in different recognition tasks with a relatively high accuracy \cite{lecun2015deep}. In this paper, we introduce a concept of augmented spikes to bridge the gaps between the spiking and non-spiking agents. In each augmented spike, both timing and spike coefficient could be used for computation and information transmission. 
The characteristics of precise analog values from ANNs and all-or-nothing spikes from SNNs are thus combined.
Our augmented spikes resemble the phenomena of burst that is widely observed in nervous systems \cite{beurrier1999subthalamic,zeldenrust2018spike,naud2018sparse,simonnet2019burst,divakaruni2018long}.

A new spiking neuron model is proposed to process augmented spikes. Notably, our scheme is not limited to a specific type of neuron model, and thus it can be easily applied to any other spiking neuron models, such as Hodgkin-Huxley model \cite{hodgkin1952quantitative}, Izhikevich model \cite{izhikevich2003simple} and hardware-friendly ones \cite{nawrocki2016mini}. 
Our results highlight that the augmented spiking neuron can not memorize more patterns than a normal one (Fig.~\ref{Fig:cpt_ac}), but it is superior to take advantage of additional information carried by spike coefficients to facilitate learning and processing.

In addition to the neuron model, several new synaptic learning rules are developed to learn from augmented spikes. Our learning rules are effective to train neurons to have binary response of firing or not, to fire at desired times and to elicit a target output spike number. Again, the idea of our augmented learning could be easily applied to other spike-based plasticity rules.

Remarkably, we successfully apply our augmented framework to two representative practical tasks: acoustic and visual pattern recognition. Spike coefficients are utilized to carry additional information that can be further used to improve the performance. These two practical tasks prove the effectiveness of our augmented encoding and learning, and they also highlight the potential merit of our approaches. In addition to these two examples, our augmented spikes could be widely used in different scenarios. In a conceivable example, a person can speak the same sentence with different emotions. Our augmented spikes could facilitate the information representation with spike timings carrying the sentence while coefficients reflecting emotions.

Our augmented spike framework is versatile, and could be generalized to many spike-based computations and plasticity schemes \cite{wu2017spiking,liu2010neuromorphic,benjamin2014neurogrid,merolla2014million,dan2004spike}. For example, in a neuromorphic sensory system \cite{liu2010neuromorphic}, the spike coefficients of our augmented spikes could be easily packed in the address-event representation (AER) and transmitted to downstream processing units.

\section{Conclusion}
\label{sec:Conclusion}
In this paper, we introduced a concept of augmented spikes to carry complementary information with spike coefficients in addition to spike latencies. We developed a new augmented spiking neuron model to process these advanced spikes. Moreover, new synaptic learning rules were proposed to train neurons to learn from augmented spikes. Therefore, we presented a new framework with augmented spikes, including focuses of coding, processing and learning. We provided systematic insight into the properties and characteristics of our proposed methods. Remarkably, our applied developments on practical recognition tasks have shown potential merits of our methods. Notably, our augmented spike framework is versatile and can be easily generalized to other spike-based systems, thus brings a new possible direction for neuromorphic computing.

\section*{Appendix}
\subsection{Momentum}
A momentum scheme could accelerate the learning \cite{gutig06}, and thus it was applied in our study.
The actual performed synaptic update $\Delta w$ was composed of two parts: the current modification, $\Delta w^\mathrm{current}$, that is determined by the corresponding learning rules, and a fraction of the previous applied update $\Delta w^\mathrm{previous}$. Therefore, in each error trial, the resulting synaptic update is as
\begin{equation}
\label{Eq:momentum}
\Delta w = \Delta w^\mathrm{current} + \mu \Delta w^\mathrm{previous}
\end{equation}
where $\mu \in [0, 1]$ is the momentum parameter determining the fraction of the previous update.

\subsection{Neuronal Parameters}
The neuronal parameters used in this paper are summarized in Table~\ref{tab:neuron}. 
\begin{table}[!tbh]
	\centering
	\caption{Neuronal parameters used in different experiments. $w_\mathrm{ini}$ represents the initialization of weights, and $\mathcal{N}(a, b)$ denotes a Gaussian distribution with the mean of $a$ and the standard deviation of $b$. The unit of $\tau_\mathrm{m}$ and $\tau_\mathrm{s}$ is millisecond.}	
	\begin{tabular}{Sc||Sc Sc Sc Sc Sc}
		\hline
		\textit{Index}  & $\tau_\mathrm{m}$ & $\tau_\mathrm{s}$ & $\eta$ & $\mu$ & $w_\mathrm{ini}$  \\ \hline 
		
		Fig.\ref{Fig:augtmp}& 20  & 5  & $10^{-4}$ & 0.9 & $\mathcal{N}(0, 0.001)$  \\

		Fig.\ref{Fig:cpt_ac}& 10  & 5  & $10^{-4}$ & 0.99 & $\mathcal{N}(0, 0.001)$  \\
		
		Fig.\ref{Fig:psd_demo},\ref{Fig:psd_wdist}& 10  & 5  & 0.01 & 0 & $\mathcal{N}(0.01, 0.01)$  \\

		Fig.\ref{Fig:lfeature}& 20  & 5  & $10^{-4}$ & 0.9 & $\mathcal{N}(0.01, 0.01)$  \\

		Fig.\ref{Fig:robustness}& 20  & 5  & $10^{-4}$ & 0.9 & $\mathcal{N}(0, 0.001)$  \\

		\begin{tabular}{@{}c@{}}AugTmp in: \\ Table \ref{tab:mismatch}\end{tabular}& 40  & 10  & $3\times10^{-3}$ & 0 & $\mathcal{N}(0, 0.01)$  \\

		\begin{tabular}{@{}c@{}}AugTDP in: \\ Fig.\ref{Fig:sound_nd},Table \ref{tab:mismatch}\end{tabular}
		& 40  & 10  & $2.5\times10^{-5}$ & 0.7 & $\mathcal{N}(0, 0.01)$  \\
		
		Table \ref{tab:mnist}
		& 40  & 10  & $2\times10^{-4}$ & 0.9 & $\mathcal{N}(0.01, 0.01)$  \\

		\hline
	\end{tabular}
	\label{tab:neuron}
\end{table}


\ifCLASSOPTIONcaptionsoff
  \newpage
\fi

\begin{IEEEbiography}[{\includegraphics[width=1in,height=1.25in,clip,keepaspectratio]{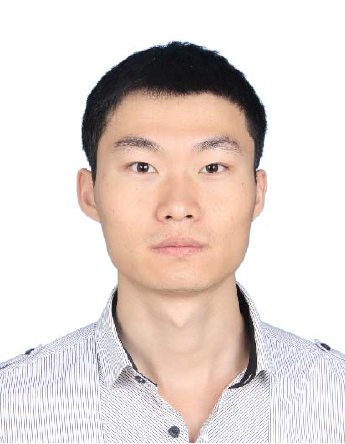}}]{Qiang~Yu}
(M'12) received the B.Eng. degree in electrical engineering and automation from the Harbin Institute of Technology, Harbin, China, in 2010, and the Ph.D. degree in electrical and computer engineering from the National University of Singapore, Singapore, in 2014.\\
He is an Associate Professor with the College of Intelligence and Computing,
Tianjin University, Tianjin, China. Before that, he was a Post-Doctoral Research Fellow with the Max-Planck-Institute for Experimental Medicine, G\"{o}ttingen, Germany, from 2014 to 2016, and a Research Scientist in the Institute for Infocomm Research, Agency for Science, Technology and Research, Singapore, from 2016. He is a recipient of the 2016 IEEE Outstanding TNNLS Paper Award. His current research interests include learning algorithms in spiking neural networks, neural coding, cognitive computations and machine learning.
\end{IEEEbiography}

\begin{IEEEbiography}[{\includegraphics[width=1in,height=1.25in,clip,keepaspectratio]{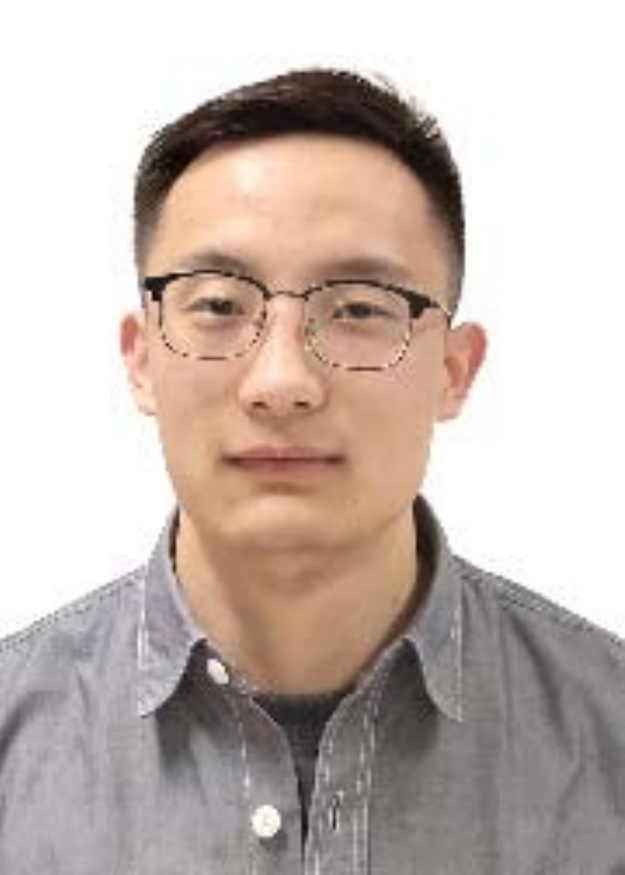}}]{Shiming~Song}
	received the bachelor's degree from Sichuan University, Chengdu, China, in 2018. He is currently pursuing the master's degree with the College of Intelligence and Computing, Tianjin University, Tianjin, China. His current research interests include spike-based learning, neural encoding and machine learning.
\end{IEEEbiography}

\begin{IEEEbiography}[{\includegraphics[width=1in,height=1.25in,clip,keepaspectratio]{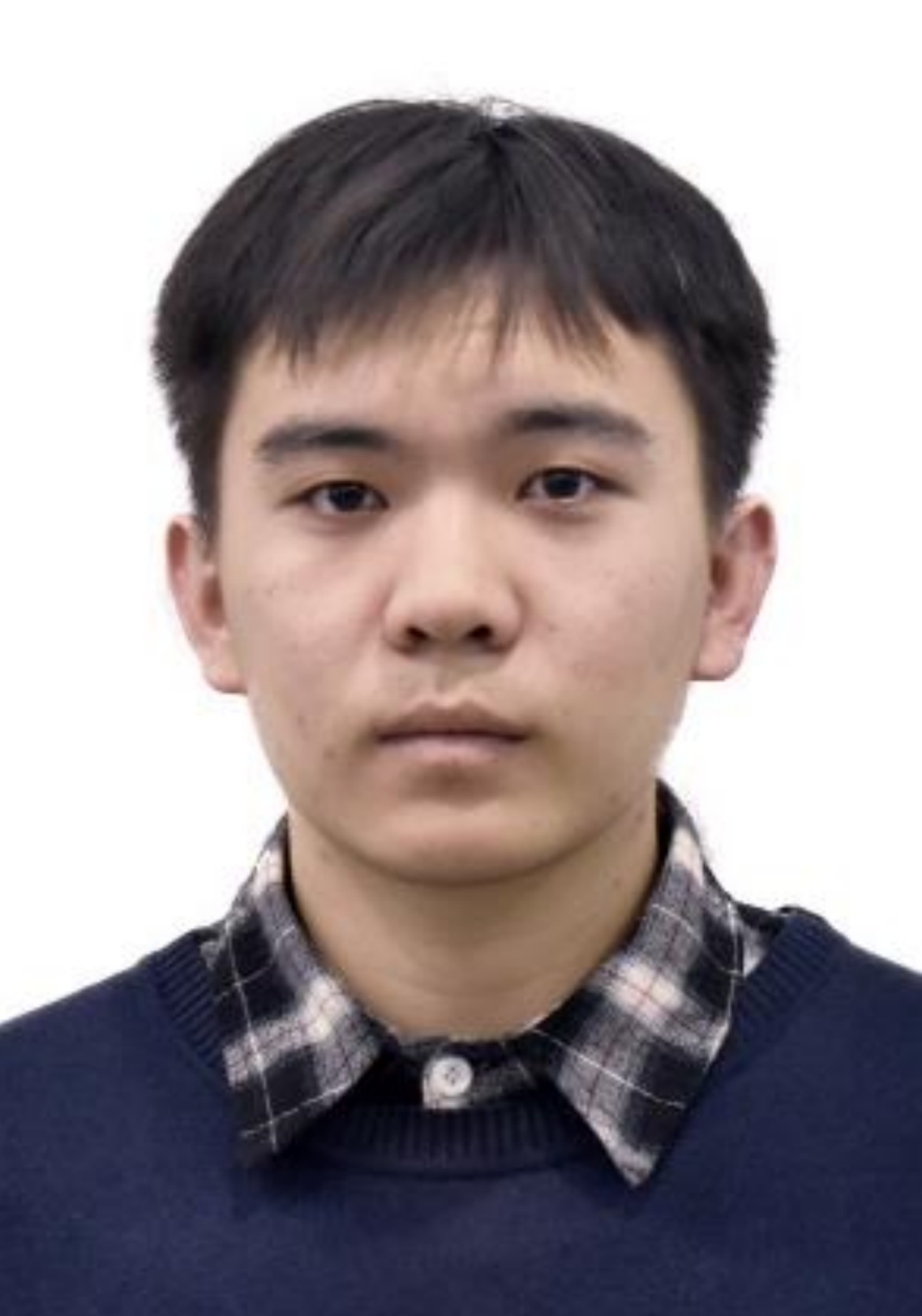}}]{Chenxiang~Ma}
	 received the B.Eng. degree from the China University of Petroleum, Qingdao, China, in 2019. He is currently pursuing the master's degree with the College of Intelligence and Computing, Tianjin University, Tianjin, China. His current research interests include learning algorithms in spiking neural network and deep learning.
\end{IEEEbiography}

\begin{IEEEbiography}[{\includegraphics[width=1in,height=1.25in,clip,keepaspectratio]{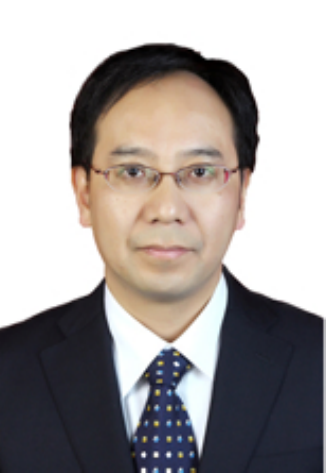}}]{Linqiang~Pan}
	(M'09) received the Ph.D. degree from Nanjing University, Nanjing, China, in 2000. He has been a Professor with the Huazhong University of Science and Technology, Wuhan, China, since 2004. His current research interests include membrane computing, DNA nanotechnology, complex network, and systems biology.
\end{IEEEbiography}

\begin{IEEEbiography}[{\includegraphics[width=1in,height=1.25in,clip,keepaspectratio]{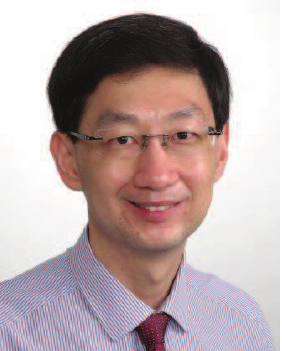}}]{Kay~Chen~TAN}
	(SM'08-F'14) received the B.Eng. (First Class Hons.) degree in electronics and	electrical engineering and the Ph.D. degree from the University of Glasgow, Glasgow, U.K., in 1994 and 1997, respectively.\\
	He is a Full Professor with the Department of Computer Science, City University of Hong Kong, Hong  Kong. He has published over 200 refereed
	articles and five books.\\
	Dr.	Tan	is the Editor-in-Chief of the IEEE TRANSACTIONS	ON EVOLUTIONARY
	COMPUTATION, was the Editor-in-Chief of the	IEEE Computational Intelligence Magazine from 2010 to 2013, and currently serves as the Editorial Board Member of over 20 journals. He is an elected member of
	the IEEE CIS AdCom from 2017 to 2019 and is an IEEE CIS Distinguished
	Lecturer from 2015 to 2017.
\end{IEEEbiography}

\end{document}